\documentclass[journal,10pt,]{IEEEtran}
\usepackage{cite}
\usepackage{amsfonts}
\usepackage[cmex10]{amsmath}
\usepackage{amssymb}
\usepackage{mathrsfs}
\usepackage[pdftex]{graphicx}
\graphicspath{{./}} 
\DeclareGraphicsExtensions{.pdf,.jpeg,.png}

\usepackage[colorlinks,linkcolor=black,urlcolor=black]{hyperref}

\usepackage{tabularx}
\usepackage{booktabs,caption,fixltx2e}
\usepackage[flushleft]{threeparttable}
\usepackage{mdwmath}
\usepackage{mdwtab}
\usepackage{array}
\usepackage{algorithm}
\usepackage{algpseudocode}
\usepackage{amsmath}
\usepackage{graphics}
\usepackage{epsfig}
\usepackage{graphicx}
\usepackage{tabularx}
\usepackage{booktabs}
\usepackage{multirow}
\usepackage{subfigure}

\IEEEoverridecommandlockouts                              
\begin{document}
\title{Robust Facial Landmark Localization Based on Texture and Pose Correlated Initialization}
\author{Yiyun Pan, Junwei Zhou, \IEEEmembership{Member, IEEE}, Yongsheng Gao,\IEEEmembership{ Senior Member, IEEE}, Shengwu Xiong
	\thanks{Y. Pan, J. Zhou and S. Xiong are with the School of Computer Science and Technology,  Wuhan University of Technology, School of Computer Science and Technology 122 Luoshi Road, Wuhan, Hubei, P.R.China, Wu Han, Hu Bei, CN 430070 , E-mail: (junweizhou@msn.com).}
	\thanks{Y. Gao is with School of Engineering, Griffith University, Australia.}
}


\maketitle

\begin{abstract}
Robust facial landmark localization remains a challenging task when faces are partially occluded. Recently, the cascaded pose regression has attracted increasing attentions, due to it's superior performance in facial landmark localization and occlusion detection. However, such an approach is sensitive to initialization, where an improper initialization can severly degrade the performance. In this paper, we propose a Robust Initialization for Cascaded Pose Regression (RICPR) by providing texture and pose correlated initial shapes for the testing face. By examining the correlation of local binary patterns histograms between the testing face and the training faces, the shapes of the training faces that are most correlated with the testing face are selected as the texture correlated initialization. To make the initialization more robust to various poses, we estimate the rough pose of the testing face according to five fiducial landmarks located by multi-task cascaded convolutional networks. Then the pose correlated initial shapes are constructed by the mean face's shape and the rough testing face pose. Finally, the texture correlated and the pose correlated initial shapes are joined together as the robust initialization. We evaluate RICPR on the challenging dataset of COFW. The experimental results demonstrate that the proposed scheme achieves better performances than the state-of-the-art methods in facial landmark localization and occlusion detection.
\end{abstract}


\IEEEpeerreviewmaketitle
\section{Introduction}
\label{sec:1}
Facial landmark localization, which is localizing the facial key points (e.g., eye brows, eyes, nose, mouth and jaw), plays an important role in many computer vision tasks, such as face detection \cite{Yang_2015_ICCV}, face recognition \cite{7377089,Li2015Towards,7448432} and facial expression analysis \cite{6482635,Kamarol201725,Zhang20153191}. In recent years, facial landmark localization has been extensively studied and achieved remarkable performance on standard datasets and even on datasets collected in the wild \cite{6248014,6909614,Zhu_2015_CVPR,6751298,7780823,7911334}. However, it still has obstacles for faces with various variations in appearance including pose, expression, especially occlusions.
\begin{figure}[!tp]
	\centering
	\includegraphics[width=3.45in]{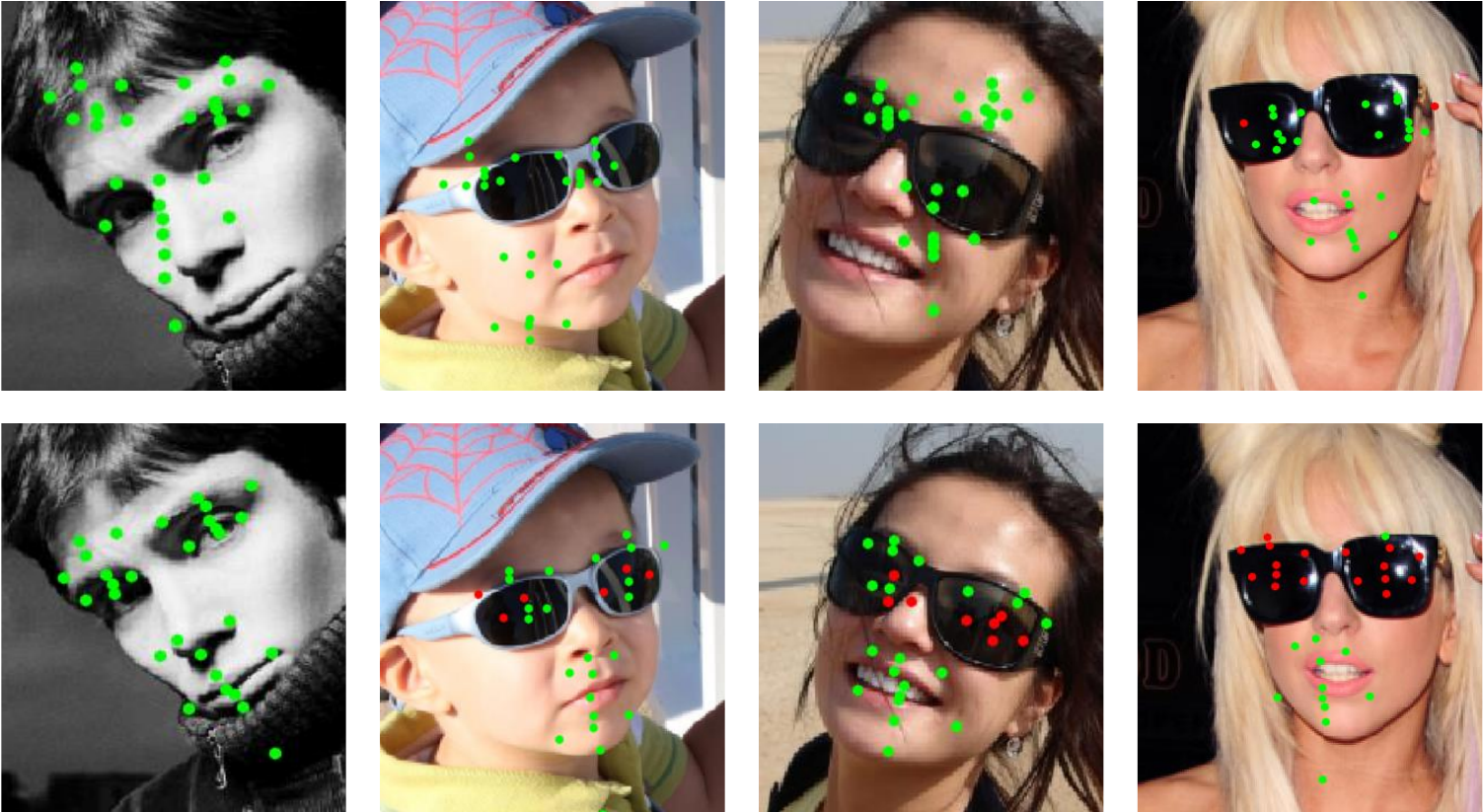}
	\centering
	\caption{Visual results of RCPR on COFW dataset (red: occluded, green: un-occluded). The initial shapes (the first row) and their localization results (the second row) of RCPR \cite{6751298}. Facial landmark localization usually fails when it begins with a bad initial shape.}
	\label{bad_initial}
\end{figure} 
\par 
Since Cascaded Pose Regression (CPR) was used to estimate facial shapes \cite{5540094}, the shape regression in a cascaded manner has emerged as one of the most popular approaches for facial landmark localization \cite{6248015,6751298,6909614,6909637,7084187,7298989,7332780,7762938,7299048,DBLP:journals/corr/SmithD16}. CPR and its variations typically begin with an initial shape, such as an average shape or a random shape of training samples, and then update the shape from coarse to fine through the trained regressors. Based on CPR, Burgos-Artizzu et al. proposed a scheme of Robust CPR (RCPR) \cite{6751298}, which is the first scheme explicitly detect occlusion state at the same time to estimate locations of landmarks. And they created a popular challenging dataset named Caltech Occluded Faces in the Wild (COFW) \cite{6751298}, where most faces in this dataset have occlusions. Researchers have used this dataset to study facial landmark localization under occlusions \cite{6751298,7084187,7762938,7780742,Yu2014,7346480}. Although these methods make some progress on facial landmark localization under partial occlusion, the occlusion problem is not essentially solved. The accuracy of occlusion prediction is still unsatisfactory. Since the occluded landmarks can hardly provide information for further analysis, it is significant to detect occlusion state of landmarks, furthermore, the occluded landmarks may reduce the accuracy of localizing the un-occluded landmarks.
\par
Since regression is initialization dependent where an improper initialization will significantly decrease the performance sharply. When the pose variation and occlusion appear simultaneously on a face, localization will fail if a bad initial shape is selected. As shown in Fig. \ref{bad_initial}, a bad initial shape usually leads to a failure of landmark localization and occlusion prediction. In this paper, we propose a Robust Initialization for CPR (RICPR)\footnote{The source code of the proposed scheme can be found at https://github.com/pervadepyy/robust-initialization-rcpr
} to avoid the bad initialization by examining texture and pose of testing face to get the texture correlated and the pose correlated initial shapes. Since texture is always related with occlusion, we select the initial shapes according to the texture correlation between the testing face and the training faces. We firstly compute Local Binary Patterns (LBP) histograms of all training faces and the testing face. Then we get correlation distance between histograms of the testing face and each training face. We choose the shapes of the most correlated training faces as the texture correlated initialization. On the other hand, the rough pose, which is represented by the rotation vector, of the testing face is used to obtain the pose correlated initial shapes. More specifically, we first estimate five fiducial landmarks including the pupils, the tip of the nose, and the corners of the mouth using Multi-Task Cascaded Convolutional Networks (MTCNN) \cite{7553523}. According to the five landmarks and a mean 3D face shape with 5 facial key points, the face pose values can be obtained. Then another 3D mean face shape, represented by 29 facial key points, can be projected to a set of corresponding 2D locations by the face pose, which can obtain the pose correlated initial shapes. Finally, the texture correlated initial shapes and the pose correlated initial shapes are taken together as the robust initialization for regression, as shown in Fig. \ref{robust_initial}, which is more relevant to the true shape of the testing face in location and occlusion. We evaluate RICPR on the challenging dataset of COFW. The experimental results show that the Normalized Mean Error (NME) is 6.64$\times10^{-2}$ and the accuracy of occlusion detection is 80/54.6\% precision/recall, which is better than that of the state-of-the-art schemes.
\par
Some of the ideas presented in this paper were initially reported in \cite{7961799}. In this paper, we report the full and new formulation and extensive experimental evaluation of our method. The initialization not only depends on texture correlation but also pose correlation and the accuracies of landmark localization and occlusion detection are further improved.
\par
The remainder of this paper are organized as follows. In Section \ref{sec:2}, the related works are briefly introduced. We review CPR and describe the proposed scheme in Section \ref{sec:3}. The experimental results on COFW dataset are given in Section \ref{sec:4}. Finally, conclusions are drawn in Section \ref{sec:5}.

\section{Related Work}
\label{sec:2}
The works to solve the problems of facial landmark localization and face alignment can be roughly divided into two groups: holistic based methods and local based methods. The holistic methods regard the shape as a whole, which usually align the face in an iterative or cascaded way. A typical holistic based method is the Active Appearance Model (AAM) \cite{Cootes1998,Matthews2004Active,Alabort-i-Medina_2014_CVPR,Antonakos2015Feature}. CPR \cite{5540094} is a similar method with a random fern regressor, which is a fast and accurate solution of computing the 2D shape of an object. Explicit Shape Regression (ESR) \cite{6248015} and RCPR \cite{6751298} extended the idea of CPR, which also use pixel difference features and fern regressor. A similar method called Supervised Descent Method (SDM) was proposed in \cite{Xiong_2013_CVPR}. This method used cascade regression with fast SIFT feature and solved localization using newton-type optimization on nonlinear least squares problem.
\par
Since occlusions are very common in the real applications of computer vision and the occluded landmarks usually cannot provide information, some works focus on facial landmark localization and occlusion detection jointly \cite{6751298,7084187,Ghiasi_2014_CVPR,7762938,Yu2014}. Burgos-Artizzu et al. first proposed to detect occlusions state at the same time of estimating landmarks in RCPR \cite{6751298}, where occlusion states are applied at each iteration to get visually different regressors. The outputs of regressors are merged with weights that depend on the occlusion prediction results. Considering occlusions often cover a region, instead of visibility annotation, Yang et, al. \cite{7084187} used the consistency of votes of the local regression forest in several over-segmented regions to get a confidence value of each pixel, which is called Regional Predictive Power (RPP). Compared with RCPR, RPP obtained a higher accuracy in landmark localization. Yu et al. \cite{Yu2014} proposed a Consensus of occlusion-robust Regression method (CoR) by forming a consensus from estimates arising from a set of occlusion-specific regressors. Each regressor is trained to estimate facial landmark locations under the precondition that a particular predefined region of the occluded face. CoR improved the accuracy of occlusion detection. Liu et, al. \cite{7762938} proposed Cascade Regression with Adaptive Shape Model (CRASM) for robust facial landmark localization. In each iteration, the shape-indexed appearance is used to estimate the occlusion level of each landmark, and each landmark is then weighted according to the occlusion level. Moreover, the occlusion level of the landmark acts as adaptive weights on the shape-indexed features to decrease the noise on the shape-indexed features, which improved the performance of facial landmark localization and occlusion detection. CRASM improved the performance of landmark localization and occlusion detection compared with other methods, which obtained a 80/48.45\% precision/recall for occlusion detection and NME is 6.68$\times10^{-2}$ for localization on COWF dataset.

\section{The Proposed Scheme}

\begin{figure*}[!tp]
	\centering
	\includegraphics[width=.99\linewidth]{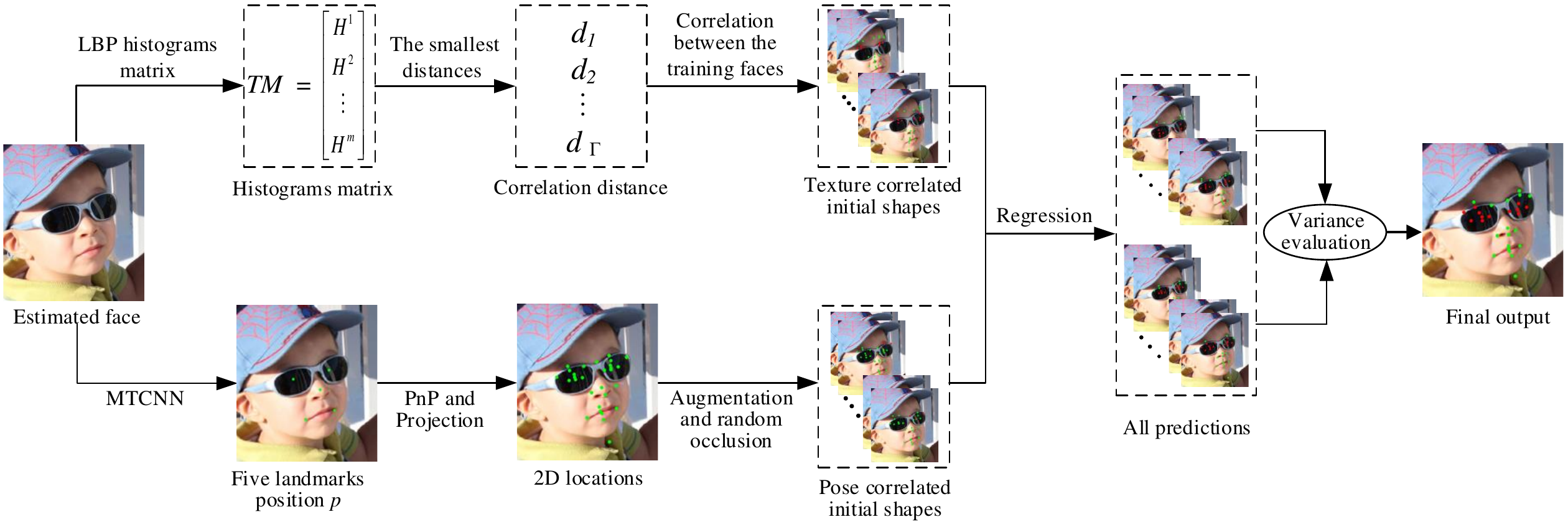}
	\caption{The procedure of RICPR. The texture correlated initial shapes and the pose correlated initial shapes are calculated in parallel. The texture correlated initialization is based on the correlation of LBP histograms between the testing face and the training faces, while the pose correlated initialization is based on the evaluated rough face pose. These initial shapes are combined together as robust initializations for regression to get predictions. Finally, the reliability of each prediction is evaluated by variance to get the final output.}
	\label{robust_initial}
\end{figure*}

\label{sec:3}
In this section, we briefly review CPR and RCPR, and then describe in detail the proposed RICPR scheme for facial landmark localization under occlusions.
 
\subsection{Cascaded Shape Regression}

\begin{algorithm}[!htbp]
	\caption{Cascaded Pose Regression}
	\label{cpr}
	\begin{algorithmic}[1]
		\Require
		Image $I$, initial shape $S^0$, regressors $R^{1..T}$ 
		\For{$t=1$ to $T$}
		\State $f^t=h^t(I,S^{t-1})$ \qquad //Shape-indexed features
		\State $\Delta{S^t}=R^t(f^t)$ \qquad //Apply regressor $R^t$
		\State $S^t=S^{t-1}+\Delta{S^t}$\qquad //Update estimated shape
		\EndFor
		\Ensure
		Estimated shape $S^T$
	\end{algorithmic}
\end{algorithm}

The main steps of CPR \cite{5540094} can be described as Algorithm \ref{cpr}. It starts from a raw initial shape $S^0$, which is progressively refined in each iteration by applying a cascade of $T$ regressors $R^{t},t=1,..,T$ until the final shape $S^T$ is estimated. At each iteration, image features are calculated as $f^t=h^t(I,S^{t-1})$, where $I$ is the face image and $S^{t-1}$ is the previous iteration's shape. Based on the shape-indexed features and the regressor $R^t$, an update $\Delta{ S^t}$ is calculated. The update $\Delta{S^t}$ is combined with $S^{t-1}$ to form a new shape $S^t$. ESR \cite{6248015} proposed some improvements over CPR, which uses two-level cascaded regression to strengthen regressors. There are $K$ primitive fern regressors $R^t=(R^t_1,..,R^t_k,..,R^t_K)$ at each iteration, and the shape update $\Delta{S^t}$ is obtained by:
\begin{equation}
\label{eqn_1}
\Delta{S^t}= \sum_{k=1}^{K} \Delta{S_k^t}=\sum_{k=1}^{K} R^t_k(h^t(I,S^{t-1})).
\end{equation}
\par
Burgos-Artizzu et al. \cite{6751298} proposed a novel regression approach RCPR, to handle localization under occlusions, which divides the face image into 9 zones. At each iteration $t$, the occlusions presented in each one of the 9 zones can be estimated by projecting the current estimation $S^{t-1}$ in the image. Then, instead of training a single regressor, RCPR trains $\eta$ regressors in each primitive fern regressor $R^t_k$. Moreover, each regressor is allowed to draw features only from 1 of the 9 predefined zones. Finally, the updates of the regressors $\delta{S_{i},i=1,..,\eta}$ are combined through a weighted mean voting. The weight $w_i^k$ is inversely proportional to the occlusions presented in the zone. At the \emph{t}-th iteration, the \emph{k}-th update can be described as:
\begin{equation}
\label{eqn_2}
\Delta{S^t_k}= \sum_{i=1}^{\eta} w_i^k \delta{S_i^k},
\end{equation}
\subsection{Robust Initialization for Cascaded Pose Regression}
The procedure of the proposed RICPR is illustrated in Fig. \ref{robust_initial}. Firstly, we get the texture correlated initial shapes by calculating texture correlation between the testing face and the training faces, at the same time, the pose correlated initial shapes are obtained by examining rough face pose of the testing face. Then, these initial shapes are taken as the robust initialization for cascaded regression. We describe these two initialization methods in the subsections \ref{lbp-i} and \ref{pose-i}, respectively. 

\subsection{ The Texture Correlated Initial Shapes }
\label{lbp-i}
Since occlusion and pose variation change the appearance of a face and texture descriptor captures the local appearance detail, we can select a texture correlated initial shape to consider the occlusion information of the testing face, rather than a random initial shape.
\par
\begin{figure}[!htbp]
	\centering
	\includegraphics[width=3.45in]{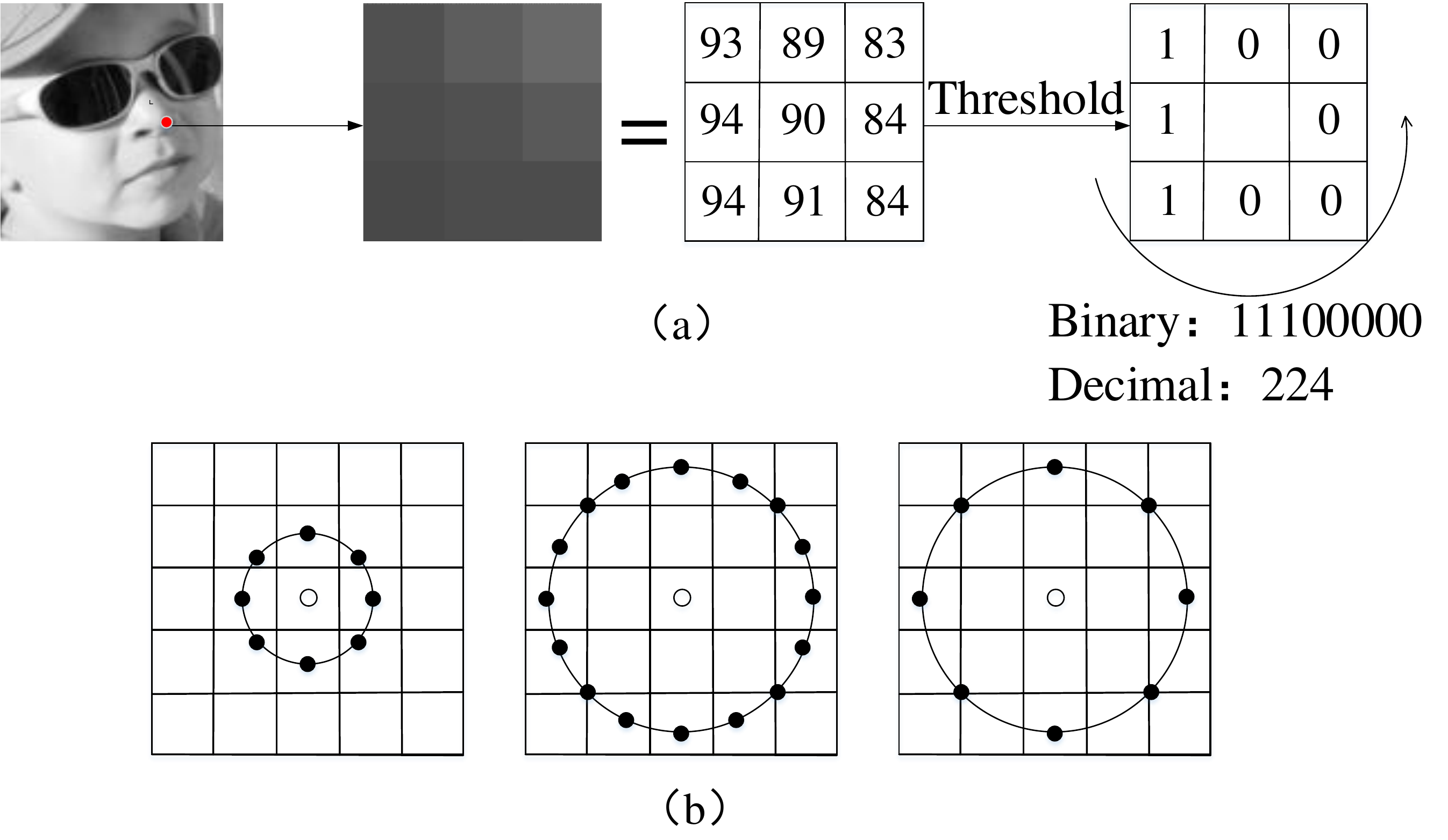}
	\centering
	\caption{(a) Basic LBP operator. (b) Examples of extended LBP operators: The circular (8,1), (16,2), and (8,2) neighborhoods.}
	\label{lbp}
\end{figure}

We propose an initialization method based on texture correlation analysis between the testing face and the training faces. The shapes of the training faces which are most correlated with the testing face are chosen as the initialization, instead of a random one. The LBP operator was originally proposed for texture analysis which is widely used in computer vision \cite{OJALA199651}. It labels an image by thresholding the $3\times 3$-neighborhood of each pixel with the centre value, as shown in Fig. \ref{lbp} (a). The histogram of the labels can be used as a texture descriptor. To deal with the limitation of the basic LBP operator, the rotation-invariant LBP and uniform LBP were proposed \cite{1017623}, as shown in Fig. \ref{lbp} (b). In the proposed scheme, we choose the uniform LBP since it balances performance and speed. We use the notation $LBP^{u2}_{P,Q}$ to denote the LBP operator, where the subscript represents using the operator in a $(P,Q)$ neighborhood, while $(P,Q)$ means $P$ sampling points on a circle of radius $Q$. The superscript $u2$ means using only uniform patterns.

\begin{figure}[!htbp]
	\centering
	\includegraphics[width=3.45in]{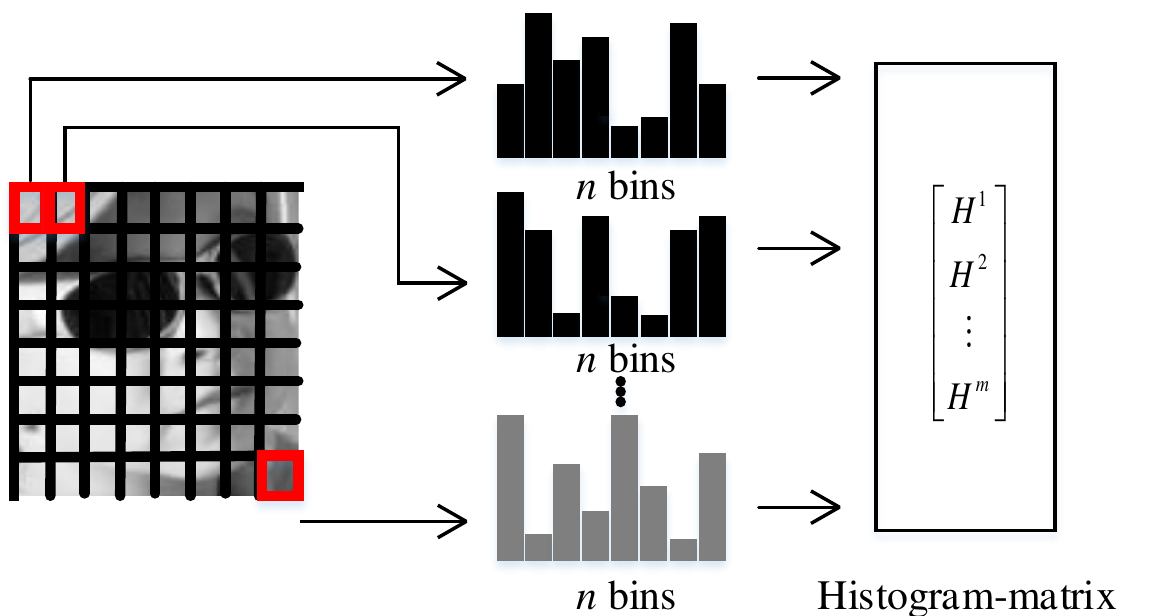}
	\centering
	\caption{The procedure of constructing histograms matrix.}
	\label{our_lbp}
\end{figure}
\par
Given an image, we divide the face into $m$ non-overlapping sub-blocks, as shown in Fig. \ref{our_lbp}. For each block $i$, we use $LBP^{u2}_{P,Q}$ to calculate LBP features, then a histogram of the labeled block $f_l(x,y)$ is computed as:
\begin{equation}
\label{eqn_3}
\begin{split}
H^i_j=\sum_{x,y} I\{f_l(x,y)=j\},I\{(x,y)\in Q_i\},&\\j=0,..,n-1,i=1,..,m,&
\end{split}
\end{equation}
where $n$ is the number of labels of each block produced by the LBP operator and $I\{ \cdot \}$ is defined as:
\begin{equation*}
\label{eqn_4}
I\{\emph{A}\}=
\begin{cases}
1,& \text{\emph{A} is true};\\
0,& \text{\emph{A} is false}.
\end{cases}
\end{equation*}
Finally, the $m$ histograms are combined yielding the histogram-matrix $H$ with a size of $m\times n$.
\par

\begin{figure*}[!tp]
	\centering
	\includegraphics[width=.90\linewidth]{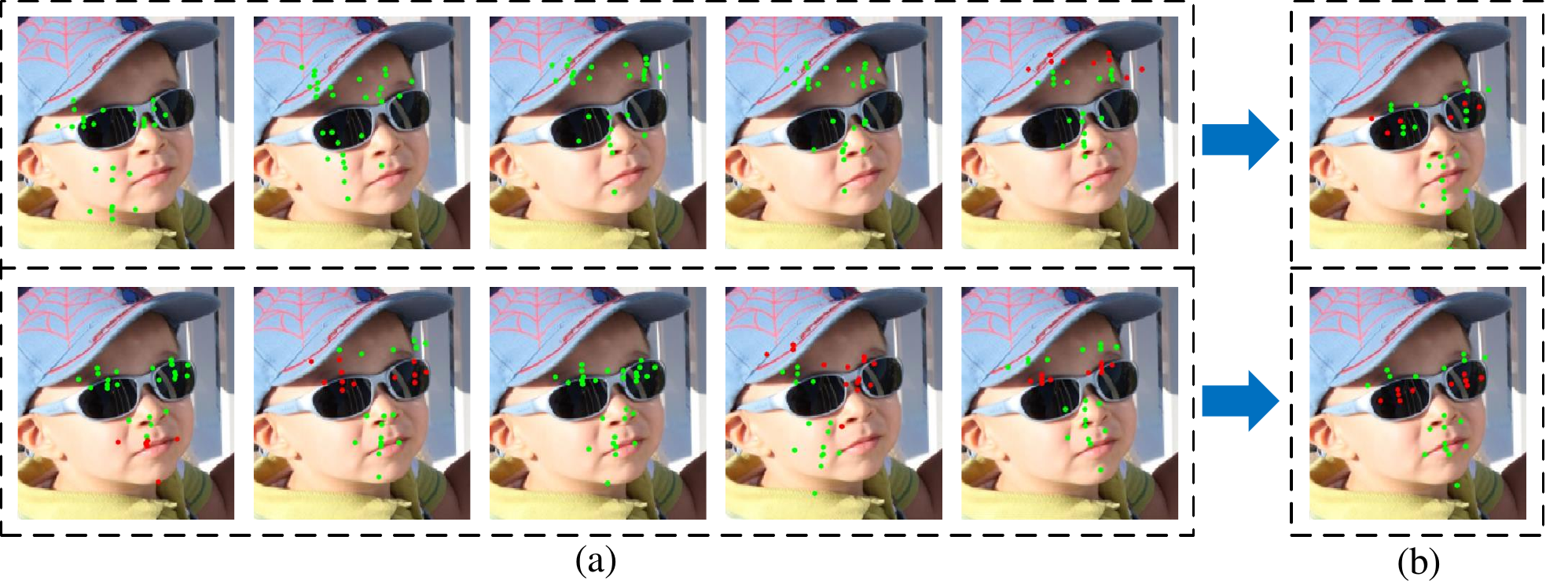}
	\centering
	\caption{We run RCPR based on 5 initial shapes selected randomly from the training set and 5 most correlated shapes from the training set by the proposed texture correlated initialization. The median of all predictions is taken as the final output. (a) The five initial shapes, where the images in the first row are the initial shapes using random initialization and the images in the second row are initial shapes using the texture correlated initialization. (b) The corresponding outputs of the two facial landmark localization methods.}
	\label{initial_compare}
\end{figure*}

During testing, histogram-matrices of the testing image and the training samples are computed by using the above scheme. It should be noticed that, to save the time cost of the testing, the histogram-matrices of the training samples can be computed offline before the testing. The best way to classify histogram-matrices is to use one of the histogram similarity measures, such as histogram intersection, log-likelihood or Chi-Square statics \cite{Ahonen2004}. Since our work aims to select proper initial shapes for regression and we hope to pick a few of the most relevant shapes with the testing face from training faces, we need a method to assess the correlation between the testing face and the training faces. In this paper, we choose the Pearson correlation coefficient \cite{10.2307/115794} to measure the correlation between the testing face and the training faces. The Pearson correlation coefficient between the testing face histogram-matrix $TM$ and each training face histogram-matrix $TRM^\gamma$ is calculated by:

\begin{algorithm}[!tp]
	\caption{Initialization based on texture correlation }
	\label{initial}
	\begin{algorithmic}[1]
		\Require
		testing face $TI$, training faces $\{TRI\}^{1..\Gamma}$, shapes of training faces $\{TRS\}^{1..\Gamma}$
		\Ensure
		Texture correlated initial shapes $\{TS_0\}^{1..l}$
		\State Compute LBP histograms for all faces
		\State Obtain histograms-matrices $TM$ and $\{TRM\}^{1..\Gamma}$ for $TI$ and $\{TRI\}^{1..\Gamma}$
		\For{$\gamma=1$ to $\Gamma$}
		\State Calculate correlation coefficient $\rho_\gamma$ between $TM$ and $TRM^\gamma$
		\State Get correlation distance $d_\gamma$
		\EndFor
		\State Search the $l$ smallest correlation distances
		\State Select corresponding $l$ shapes in $\{TRS\}^{1..\Gamma}$ as $\{TS_0\}^{1..l}$
	\end{algorithmic}
\end{algorithm}

\begin{equation}
\label{eqn_5}
\begin{aligned}
\rho_\gamma&=\frac{Cov(TM,TRM^\gamma)} {\sigma{(TM)}\sigma{(TRM^\gamma)}}\\&=\frac{E[(TM-E(TM))(TRM^\gamma-E(TRM^\gamma))]} {\sigma{(TM)}\sigma{(TRM^\gamma)}},\\&\gamma=1,..,\Gamma,
\end{aligned}
\end{equation}
where $Cov(\cdot,\cdot)$ is the covariance, $\sigma(\cdot)$ is the standard deviation and $\Gamma$ is the total number of training faces. Due to the size of each histogram-matrix is $m\times n$, $\rho_\gamma$ can be calculated as:
\begin{equation*}
\label{eqn_6}
\frac{\sum\limits_{i=1}^{m}\sum\limits_{j=1}^{n} (TM_{ij}-\overline{TM})(TRM_{ij}^\gamma-\overline{TRM^\gamma})} {\sqrt{\sum\limits_{i=1}^{m}\sum\limits_{j=1}^{n} (TM_{ij}-\overline{TM})^2} \sqrt{\sum\limits_{i=1}^{m}\sum\limits_{j=1}^{n} (TRM_{ij}^\gamma-\overline{TRM^\gamma})^2},}
\end{equation*}
where $\overline{TM}$ and $\overline{TRM^\gamma}$ are mean values of matrices ${TM}$ and ${TRM^\gamma}$, respectively. Then the correlation coefficient $\rho^\gamma$ can be used to calculate correlation distance $d_\gamma$:
\begin{equation}
\label{eqn_7}
d_\gamma=1-\rho_\gamma, \gamma=1,..,\Gamma.
\end{equation}
\par
A smaller $d$ represents that the training face is more correlated with the testing face. We choose the $l$ most correlated faces from $\Gamma$ training faces and select their shapes as initial shapes for the testing face. The main procedure of initialization based on texture correlation is presented in Algorithm \ref{initial}. As shown in Fig. \ref{initial_compare}, a comparison between random initialization \cite{6751298} and texture correlation based initialization is illustrated. It can be found that the proposed texture correlation based initialization usually obtains more accurate initial shapes which improves the accuracy of landmark localization.

\subsection{The Pose Correlated Initial Shapes}
\label{pose-i}
In the above section, we describe how to select the texture correlated initial shapes considering the occlusion information but ignoring pose information of the testing face. Empirically, landmark distribution is highly correlated to head pose. To further make the initial shapes more robust to various poses, we choose some pose correlated initial shapes for regression.
\par
\begin{figure}[!tp]
	\centering
	\includegraphics[width=.99\linewidth]{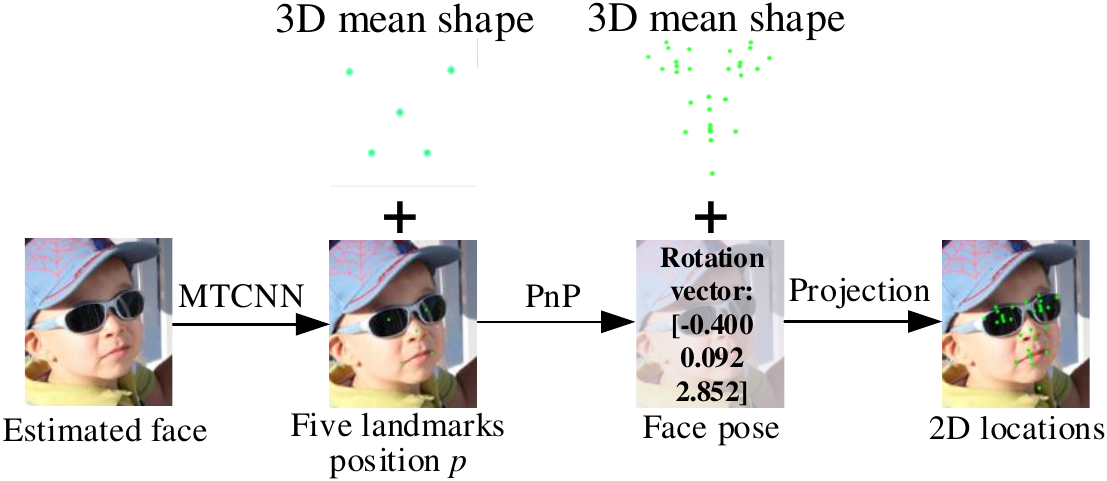}
	\caption{Illustration of generating the pose correlated shape. Given an image, we first detect five fiducial landmarks and estimate face pose. Then, according to the face pose, a 3D mean face shape with 29 facial key points, can be projected to a set of corresponding 2D locations, which has similar pose with testing image.}
	\label{pose_initial}
\end{figure}

To obtain the pose correlated initial shapes, we estimate the rough pose of the testing face, which can be obtained by the five fiducial landmarks, i.e., the pupils, the tip of the nose, and the corners of the mouth. In this paper, we use MTCNN \cite{7553523} to detect the five fiducial landmarks, as shown in Fig. \ref{pose_initial}. Inspired by Perspective-n-Point (PnP) problem, which is the problem of estimating the pose of a calibrated camera given a set of 3D points and their corresponding 2D projections in the image \cite{Lepetit2009EPnP}. Given a 3D mean shape $\overline{S}$ with 5 facial key points and the five detected fiducial landmarks, a rough face pose can be estimated by:
\begin{equation}
\label{eqn_8}
\vec{\theta} = g( \overline{S} , p),
\end{equation}
where $\vec{\theta}$ is a rotation vector, which represents the face pose and $p$ represents the five fiducial landmarks detected by MTCNN, $g$ is the Efficient PnP \cite{Lepetit2009EPnP}.
\par

\begin{figure*}[!tp]	
	\centering
	\subfigure[]{\label{error_rank}
		\begin{minipage}[b]{0.4\textwidth}
			\includegraphics[width=1\textwidth]{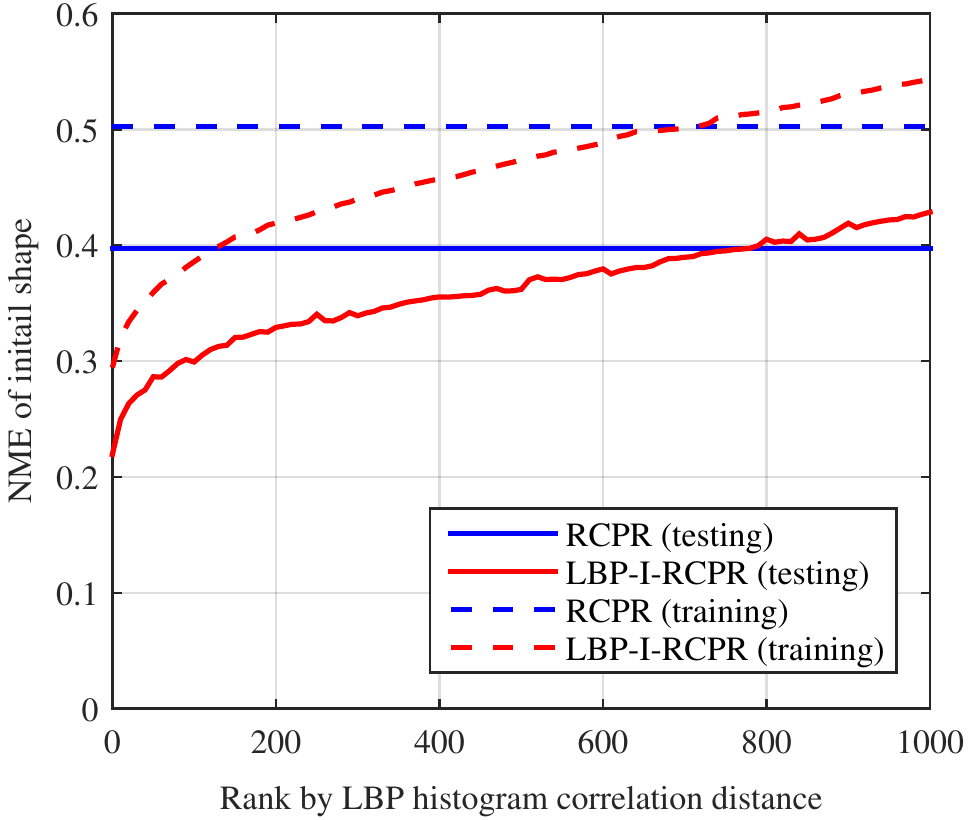} 
		\end{minipage}
	}
	\subfigure[]{\label{good_rank}
		\begin{minipage}[b]{0.4\textwidth}
			\includegraphics[width=1\textwidth]{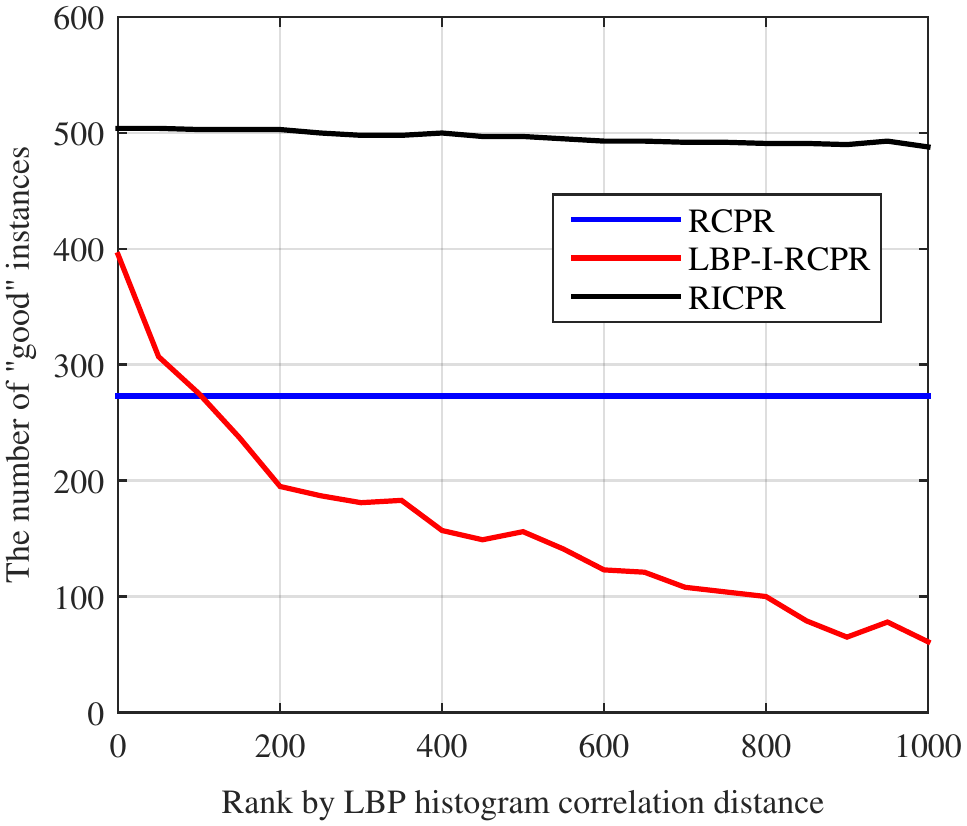} 
		\end{minipage}
	}
	\caption{Comparisons between the texture correlated initialization based RCPR and the traditional random initialization based RCPR. (a) The NME on various initial shapes with different correlation distances in training and testing processes. (b) The number of ``good'' instances determined by variance after 10\% cascades of each prediction on various correlation distances. Correlation distances in (a) and (b) are ranked in ascending order.}
	\label{ini_result}
\end{figure*}

\begin{table*}[!tp]
	\renewcommand\arraystretch{1.5}
	\begin{center}
		\normalsize
		\caption{\label{feture_compare} Texture Correlated Initialization Using Different Features }
		\begin{tabular}{|p{2.4cm}<{\centering}|p{1.7cm}<{\centering}|p{1.7cm}<{\centering}| p{1.7cm}<{\centering}|p{1.7cm}<{\centering}|p{1.7cm}<{\centering}|p{1.7cm}<{\centering}|p{1.7cm}<{\centering}|}
			\hline
			\textbf{ Feature} &\textbf{LBP} &\textbf{LDP} &\textbf{Gabor} &\textbf{GMRF} &\textbf{GLDS} &\textbf{GLCM} &\textbf{Eigenface}\\
			\hline
			\textbf{NME($\times10^{-2}$)}& 7.35 & 7.75 & 7.87 & 8.28 & 8.19 & 8.06 &8.18\\
			\hline
			\textbf{Precision/recall}& 80/51.4\% & 80/48.7\% & 80/46.1\% & 80/45.6\% & 80/47.2\% & 80/46.5\% &80/47.6\%\\
			\hline
		\end{tabular}
	\end{center}
	\begin{tablenotes}
		\footnotesize
		\item Accuracy of facial landmark localization and occlusion detection based on texture correlated initialization using different features. The results indicate that the LBP performs better than the others.
	\end{tablenotes}
\end{table*}

Then, a 3D mean face shape, represented by 29 facial landmark locations, is projected to a set of corresponding 2D locations according to the testing face pose $\vec{\theta}$, as shown in Fig. \ref{pose_initial}. After that, the shape which has similar pose with the testing face is obtained. To get a reasonable initial shape for each image, we re-scale the corresponding 2D locations based on the face bounding box and the detected five fiducial landmarks $p$. The initial occlusion information of the pose correlated initial shape is distributed randomly as:
\begin{equation}
\label{eqn_9}
S_0 = G( \overline{S^*} , \vec{\theta}, b, p),
\end{equation}
where $b$ is the face bounding box, $\overline{S^*}$ is the 3D mean face shape with 29 points, and $S_0$ is the pose correlated initial shape. To achieve a better performance, we selected several frontal faces from the training set to augment the pose correlated initial shape. Referring to their true 2D shapes and the 3D mean face shape $\overline{S^*}$, we construct different 3D frontal face shapes that have little variation compared with the 3D mean face shape. Then, based on Eq. \ref{eqn_9}, different initial shapes can be generated by replacing $\overline{S^*}$ with the constructed 3D frontal face shapes.	

\subsection{Variance Evaluation}
As stated in \cite{6751298}, due to the coarse to fine nature of CPR, even if a face image is initialized by several different shapes, the predictions should reach a similarity after iterations. Based on this principle, instead of taking the median of all predicted results as the final output, the variance is used to determine the reliability of two initialization methods' predictions. 
\par
Firstly, after finishing the regression, the variance $v$ of all predictions is calculated. If the value of $v$ is below a certain threshold $\zeta$, it indicates that the predictions is a good solution. In this case, all predictions are considered as reliable, thus we take the median of all predictions as the final output. Otherwise, part of predictions belong to ``bad'' class, then the variances between predictions based on two initialization methods are computed and represented as $v_{lbp}$ and $v_{pose}$. If $v_{lbp}$ is less than $v_{pose}$, it indicates that the predictions based on the texture correlated initialization are more reliable than these based on the pose correlated initialization. Therefore, only considering the predictions by the texture correlated initial shapes, we abandon the predictions which make obvious variance variation and take the median of the rest of predictions as the final output. If $v_{lbp}$ is greater than $v_{pose}$, it indicates that the predictions based on the pose correlated initialization are more reliable than those based on the texture correlated initialization. Then, the median of the predictions based on the pose correlated initial shapes are taken as the final output.

\section{Experimental Results }
\label{sec:4}

\subsection{Dataset and Implementation}

\begin{table}[!tp]
	\begin{center}
		\begin{threeparttable}
			
			\renewcommand\arraystretch{1.4}
			\caption{\label{err/fail}
				Results on COFW Dataset.}			
			\begin{tabular}{l | c | c }
			\hline
				\multirow{2}*{Methods} & {Landmark localization error} & {Occlusion prediction} \\
				\cline{2-3}
				 & {NME ($\times10^{-2}$)} & {Precision/Recall} \\
				\hline
				\textbf{RCPR} \cite{6751298} & 8.01 & 80/42\%\\
				\textbf{HPM} \cite{Ghiasi_2014_CVPR} & 7.46* & 80/37\%*\\
				\textbf{RPP} \cite{7084187} & 7.52* & 78/40\%*\\
				\textbf{SDM} \cite{Xiong_2013_CVPR} & 10.88 & -\\
				\textbf{TCDCN} \cite{7208848} & 8.05* & -\\
				\textbf{CRASM} \cite{7762938} & 6.68* & 80/48.45\%*\\
				\textbf{HORSD} \cite{7911334} & 6.8* & -\\			
				\textbf{LBP-I-RCPR} & 7.35 & 80/51.4\%\\
				\textbf{RICPR} & 6.64 & 80/54.6\%\\
				\textbf{Human} \cite{6751298} & 5.6 & - \\
				\hline
			\end{tabular}
			
			\begin{tablenotes}
				\footnotesize
				\item Comparison of facial landmark localization and occlusion prediction on COFW dataset. The table lists the results of NME and occlusion detection. * indicates that the result is from the published paper.
			\end{tablenotes}
		\end{threeparttable}
	\end{center}
\end{table} 

We evaluate the performance of the proposed scheme on the challenging dataset COFW \cite{6751298}, which is widely used to evaluate the robustness of facial landmark localization and occlusion detection. The face images in COFW have large variations in shape and occlusions due to differences in pose, expression, hairstyle, using of accessories such as sunglasses, hats and interactions with objects (e.g. food, hands, microphones, etc.). Each image is annotated with the location and occluded/un-occluded state of 29 facial landmarks. This dataset has 1852 face images in total, where 1345 and 507 images are used for training and testing respectively. The average occlusion rate of faces in COFW is over 23\%.
\par
To evaluate the performance of the proposed scheme, we implement the proposed scheme with two configurations. One is RCPR based on LBP histogram correlation initialization (LBP-I-RCPR), which represents the initialization only based on texture correlation, and the median of predictions is taken as the final output. The other is the full version of the proposed scheme RICPR, in which the initialization is based jointly on both texture correlation and pose correlation jointly. In texture correlation analysis, the face, whose location is provided by a face detector, is divided into $m=8\times 8=64$ non-overlapping sub-blocks. The uniform LBP in a $(P,Q)=(8,1)$ neighborhood is employed to obtain texture information of the face. Thus, the number of labels produced by the LBP operator is 59. In pose correlation analysis, we utilize MTCNN \cite{7553523} to predict five fiducial landmarks. A threshold $\zeta=0.08$ is used to determine whether the predictions lead a good result. Since the proposed scheme always has a good initialization without the need for the smart restarts and the number of initial shapes is set to 10 and $\eta$ is set to 4. We run RICPR and RCPR using the same configuration.
\par
We compare LBP-I-RCPR and RICPR with several state-of-the-art methods on COFW dataset using NME (Normalized Mean Error) defined by Eq. \ref{eqn_10}. 
\begin{equation}
\label{eqn_10}
NME=\frac{1}{N}\sum\limits_{i}^{N} \frac{\frac{1}{M}\sum\limits_{j}^{M}|s^p_{i,j}-s^g_{i,j}|_2} {|gl_i-gr_i|_2},
\end{equation}
where $N$ is the number of images in the test set, $M$ is the number of landmarks in one image, $s^p_{i,j}$ is the predicted position of the $j^{th}$ landmark of the $i^{th}$ image, $s^g_{i,j}$ is the ground truth position of the $j^{th}$ landmark of the $i^{th}$ image, $gl_i$ and $gr_i$ are the ground truth positions of the left and right eye centres respectively.
\par
Based on NME, we can plot the Cumulative Error Distribution (CED) curves to further analyse the performance of the proposed scheme, which is calculated from the NME over each image. We also evaluate the speed of the proposed scheme on the COFW dataset. Speed is measured in Frames Per Second (FPS). All methods are implemented using Matlab R2015b and run on a PC with 3.60 GHz CPU and 64-bit Windows 7 operating system. 

\begin{figure}[!tp]
	\centering
	\includegraphics[width=3.4in]{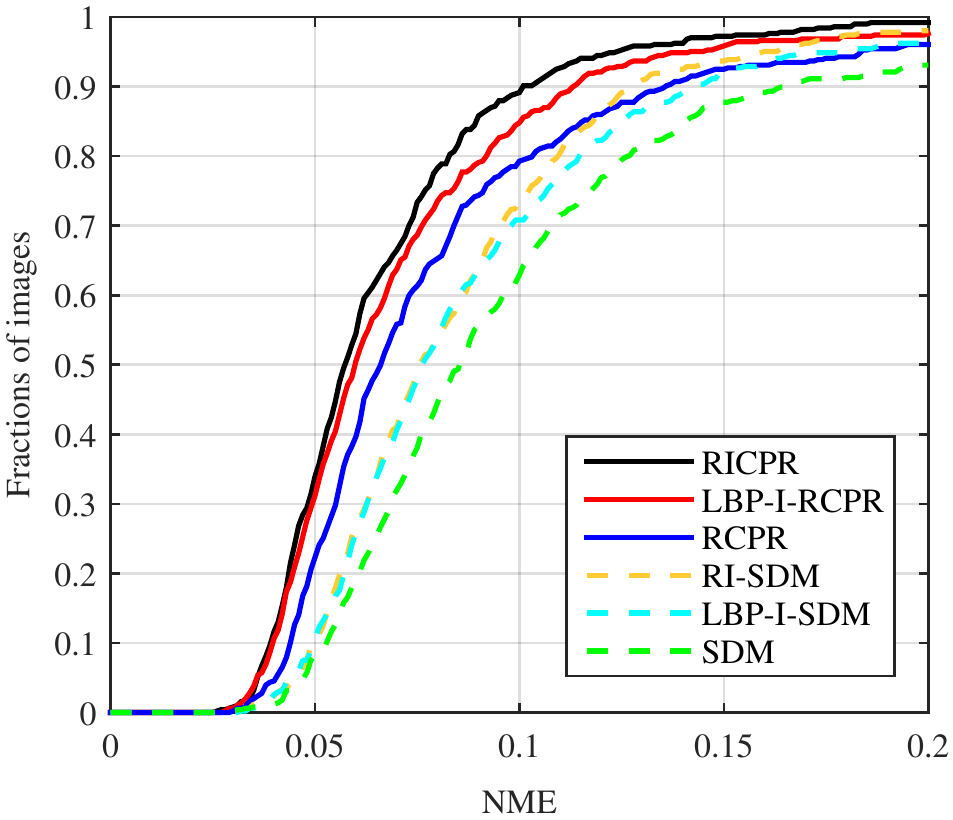}
	\centering
	\caption{CED curves on the COFW dataset.}
	\label{ced}
\end{figure}

\subsection{Results}
\textbf{1) Analysis of initialization based on texture correlation:} Instead of randomly selecting shapes from the training set as the initialization in RCPR, we employ a texture correlated initialization (LBP-I-RCPR) by computing LBP histograms. To prove the effectiveness of the texture correlated initialization method, we compare the performance of LBP-I-RCPR with RCPR on the COFW dataset as shown in Fig. \ref{ini_result}.
\par
The NMEs on various initial shapes with different correlation distances are shown in Fig. \ref{error_rank}. The results show that the NME is reduced with decreasing correlation distance and LBP-I-RCPR can significantly reduces NME by at least 45\%. It indicates that the initial shapes which are selected from the training faces based on texture correlation is closer to the real shape of the testing face.
\par
Moreover, given different initial shapes for each image, the variance between their predictions is applied to determine whether the face belongs to a ``good'' class as stated in \cite{6751298}. As shown in Fig. \ref{good_rank}, it can be found that the number of ``good'' instances increases as correlation distance decreases and more images belong to ``good'' class among 507 testing images when using LBP-I-RCPR. The number of ``good'' instances dramatically increase by at least 45\%, and thus less bad initial shapes are selected. Furthermore, the number of ``good'' instances increases from 395 to 504 among the 507 images in RICPR scheme, which means fewer than 1\% instances are ``bad'', thus the initialization become more robust.
\par
We also initialize the shapes using other different features, including Local Derivative Pattern (LDP) \cite{Zhang2010Local}, Gabor, Gaussian Markov Random Field (GMRF), Gray-Level Difference Statistics (GLDS), Gray-Level Co-occurrence Matrix (GLCM), and Eigenface. We report the NME and occlusion detection of each feature respectively in Table \ref{feture_compare}. The results indicate that the initialization based on LBP histogram correlation performs better.

\begin{figure}[!tp]
	\centering
	\includegraphics[width=3.4in]{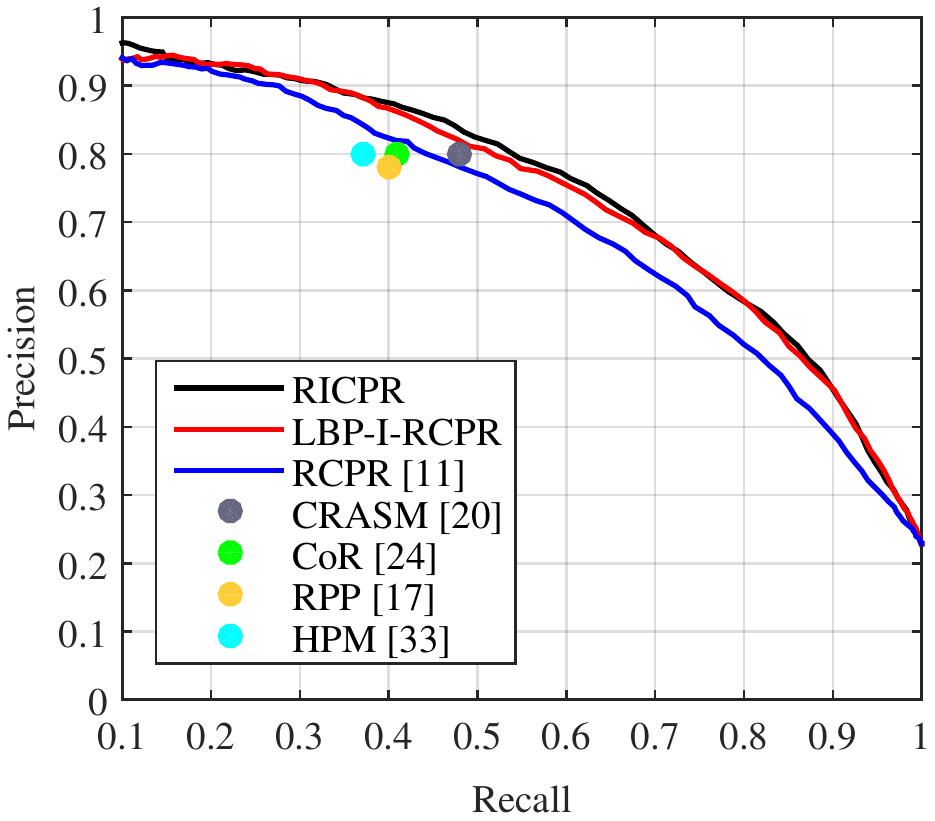}
	\centering
	\caption{Occlusion detection result on the COFW dataset.}
	\label{re/pre}
\end{figure}

\par
\textbf{2) Facial landmark localization evaluation on COFW:} Many facial landmark localization methods perform not well on the COFW database due to the large variation in occlusion. To evaluate the proposed scheme, we compare the proposed scheme with several state-of-the-art methods including RCPR \cite{6751298}, RPP \cite{7084187}, SDM \cite{Xiong_2013_CVPR}, Tasks-Constrained Deep Convolutional Network (TCDCN) \cite{7208848}, Hierarchical Deformable Part Model (HPM) \cite{Ghiasi_2014_CVPR}, CRASM \cite{7762938} and Hierarchical Occlusion Stage-wise Relational Dictionary (HOSRD)\cite{7911334}. The comparisons of NME on COFW dataset are given in Table \ref{err/fail}.
\par
We can find that RICPR obtains the smallest NME. Compared to RCPR, the LBP-I-RCPR reduces the NME from 8.01$\times10^{-2}$ to 7.35$\times10^{-2}$ and the RICPR further reduces the NME to 6.64$\times10^{-2}$. The NME is reduced by 17.1\% in total. RICPR performs even better than the most recent CRASM method proposed in 2017. To get the pose correlated initial shapes, we use MTCNN to detect five fiducial landmarks. The accuracy of five fiducial landmarks plays a significant role on performance. If the ground-truth of the five fiducial landmarks is employed, the NME can reach 5.52$\times10^{-2}$, which demonstrates that the proposed scheme can obtain a admirable performance if the five fiducial landmarks are detected accurately.
\par
We also show the CED curves of the COFW dataset in Fig. \ref{ced}. As can be seen, more images perform better using the proposed scheme, it also demonstrates the superiority of the proposed scheme for facial landmark localization in face image with occlusions.

\begin{figure}[!tp]
	\centering
	\includegraphics[width=3.4in]{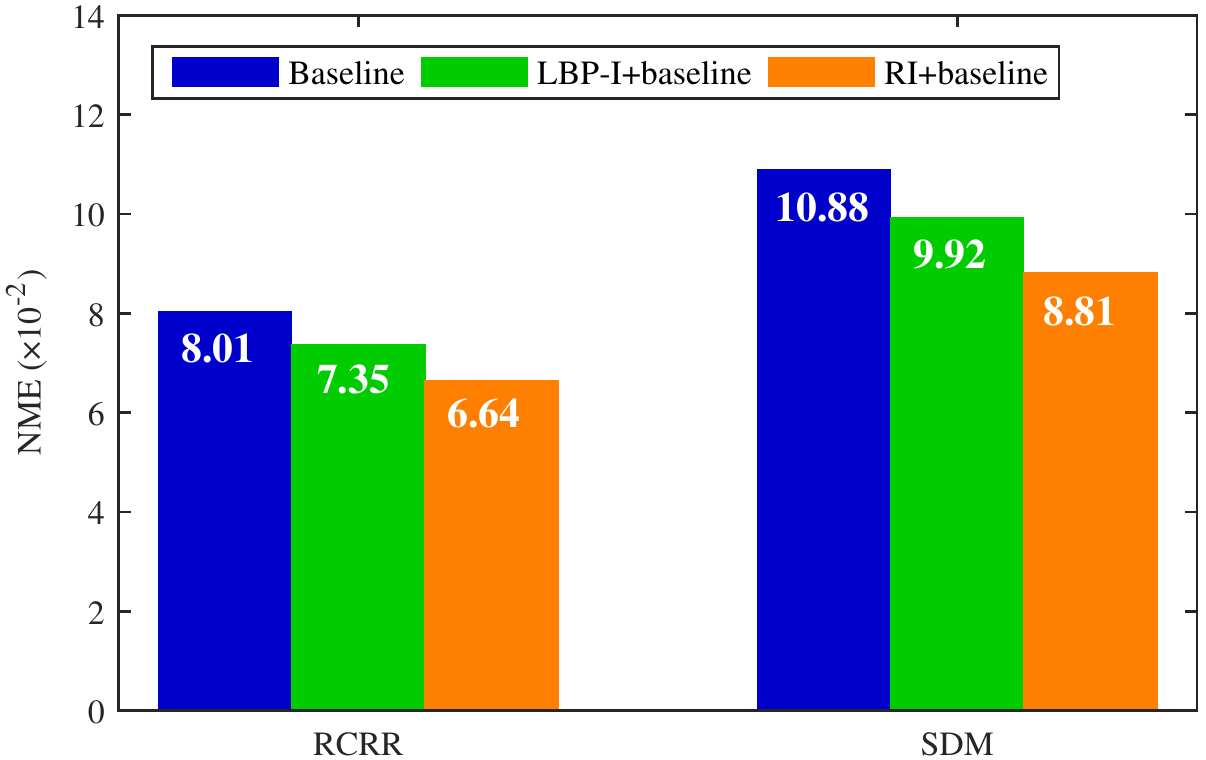}
	\centering
	\caption{Results of SDM and RCPR using the proposed initialization methods.}
	\label{sdm-rcpr}
\end{figure}

\par
\textbf{3) Occlusion detection on COFW:} Since the COFW dataset provides the ground truth of occlusion, we evaluate the occlusion detection on COFW and compare the proposed scheme with RCPR \cite{6751298}, HPM \cite{Ghiasi_2014_CVPR}, CoR \cite{Yu2014}, RPP \cite{7084187} and CRASM \cite{7762938}. The occlusion prediction results are shown in Table \ref{err/fail} and Fig. \ref{re/pre}. As can be seen, the proposed scheme also outperforms the state-of-the-art methods in occlusion detection. 
\par
\begin{figure*}[!htbp]
	\centering
	\includegraphics[width=.95\linewidth]{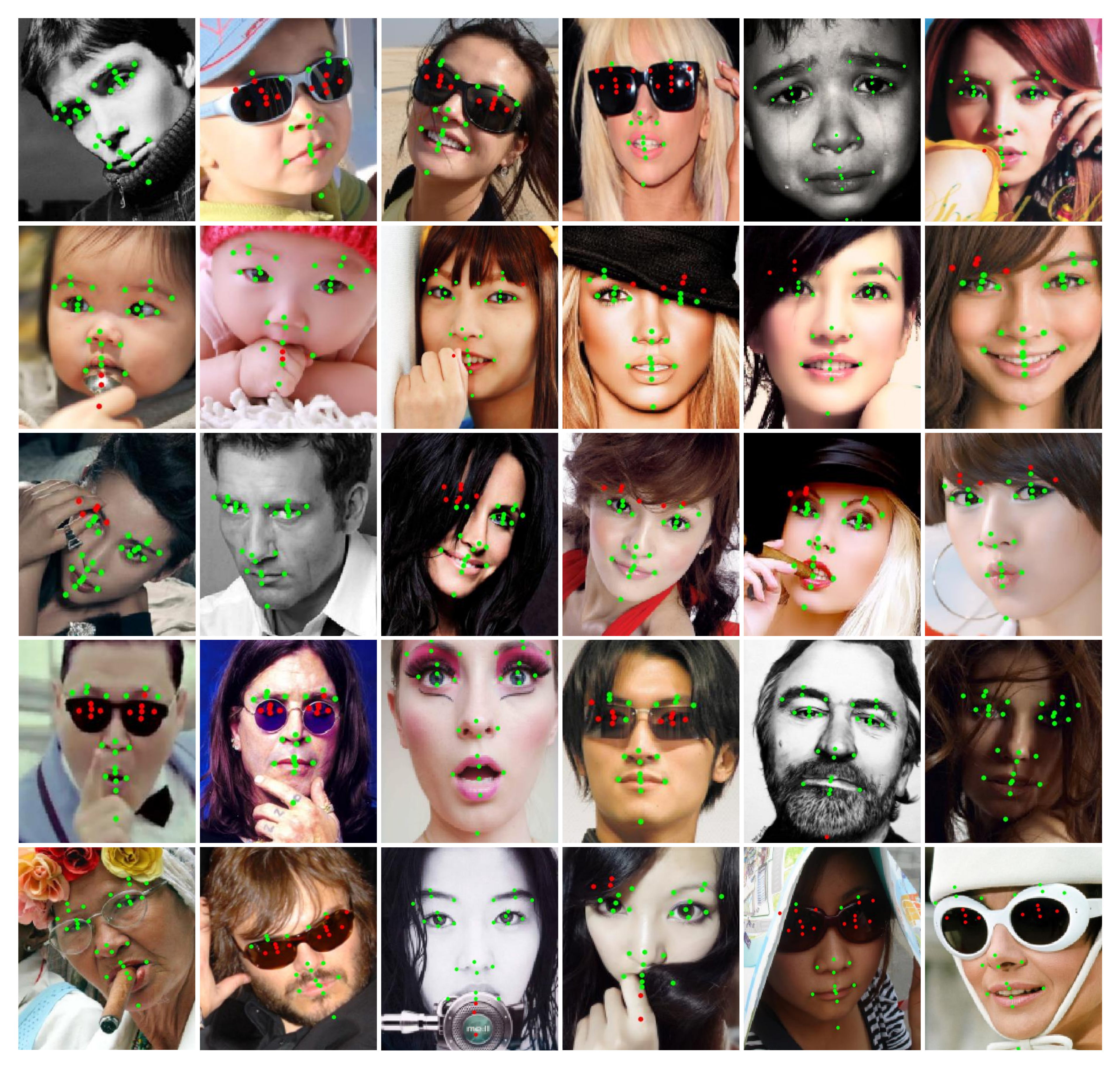}
	\caption{Example result of facial landmark localization and occlusion detection obtained by the proposed RICPR on the COFW dataset.}
	\label{sample_result} 
\end{figure*}

When we set the false alarm at 80\%, the proposed scheme achieves an accuracy of 54.6\%, which is higher than 42\% obtained by RCPR, 37\% obtained by HPM, 41.44\% obtained by CoR, 48.45\% obtained by CoR and 78/40\% precision/recall obtained by RPP. Even if only using LBP-I-RCPR scheme, the accuracy of detecting occlusion reaches 51.4\%. It demonstrates that the proposed scheme achieves a much higher accuracy of occlusion detection, which can provide significant benefits in real world application, such as image texture analysis, facial expression understanding and face recognition. Fig. \ref{sample_result} shows example images with the result obtained by the proposed RICPR.

\par
\textbf{4) Run time:} We record the speeds of RCPR, LBP-I-RCPR and RICPR on the COFW dataset. The speeds of these methods are 5.3 FPS, 4.1 FPS and 4.0 FPS, respectively. We can find that the proposed scheme takes some time on calculating the correlation. The speed can be improved by implementing it with C++ or using a powerful server. We will try to improve the efficiency of the proposed scheme in the future, for example, by reducing the number of face images used for texture correlation based initialization.

\subsection{Generalization of the Proposed Initialization Scheme}
The experimental results demonstrate that the proposed initialization scheme significantly improved the performance of RCPR in both localization and occlusion prediction. Since the initialization is usually independent to facial landmark localization, the proposed initialization scheme can be applied to other algorithms such as SDM. The results are shown in Fig.\ref{sdm-rcpr}, where baseline is the original SDM or RCPR, LBP-I+baseline is the texture correlated initialization scheme applied to SDM or RCPR, RI+baseline denotes the joint texture correlation and pose correlation initialization scheme applied to SDM or RCPR. Compared with the original SDM which is based on random initialization, the LBP-I-SDM and the RI-SDM reduce the NME by 14.6\% and 19\% respectively. The results indicate that the proposed initialization scheme can also improve the performance of SDM. 

\section{Conclusions}
\label{sec:5}
In this paper, we propose a robust initialization scheme to solve the initialization sensitive problem for the cascaded pose regression approach through jointly analyzing texture and pose of a testing face. By examining the correlation of local binary patterns histograms between the testing face and the training faces, the texture correlated shapes are selected instead of random shapes. At the same time, the pose correlated initialization is proposed to further improve the robustness of the initialization by estimating the face pose. Experimental results show that the proposed scheme obtains remarkably higher accuracies on both facial landmark localization and occlusion detection on facial images than the state-of-the-art benchmarks. Moreover, since the initialization is usually independent with facial landmark localization, the proposed initialization scheme has the potential to be extended and applied to other algorithms.

\section*{Acknowledgment}
This work is supported by the National Natural Science Foundation of China (Grant No. 61601337).

\bibliographystyle{IEEEtran}
\bibliography{DAC}
\end{document}